\documentclass{article}

\usepackage{arxiv}
\usepackage{multirow}
\usepackage{array}
\usepackage{float}
\usepackage{framed,multirow}
\usepackage{latexsym}
 \usepackage{xcolor}
\usepackage{acronym}
\usepackage{graphicx}
\usepackage{amssymb}
\usepackage{amsmath}
\usepackage{url}
\usepackage{mathtools}
\usepackage{booktabs}
\usepackage{tabularx}

\usepackage[labelformat=simple]{subcaption}

\DeclareCaptionLabelFormat{subcaptionlabel}{\normalfont(\textbf{#2}\normalfont)}
\usepackage{xcolor}

\definecolor{alexcolor}{rgb}{0.4,0.6,0.2}
\definecolor{costincolor}{rgb}{0.6,0.2,0.4}

\newcommand{\hl}[1]{#1}

\DeclarePairedDelimiterX{\infdivx}[2]{(}{)}{%
	#1\;\delimsize\|\;#2%
}

\DeclareMathOperator*{\argmax}{arg\,max}
\DeclareMathOperator*{\sg}{sg}

\newcommand{\ie}{i.e., }
\newcommand{\eg}{e.g., }
\newcommand{\wrt}{w.r.t. }

\definecolor{newcolor}{rgb}{.8,.349,.1}
\acrodef{ML}[ML]{machine learning}
\acrodef{AI}[AI]{artificial intelligence}
\acrodef{RF}[RF]{random forest}
\acrodef{DL}[DL]{deep learning}
\acrodef{NN}[NN]{neural networks}
\acrodef{DNN}[DNN]{deep neural network}
\acrodef{SSL}[SSL]{Self-Supervised Learning}
\acrodef{OOD}[OOD]{Out-of-Distribution}
\acrodef{VAE}[VAE]{Variational Autoencoder}
\acrodef{DRL}[DRL]{Disentangled Representation Learning}
\acrodef{EMA}[EMA]{Exponential Moving Average}
\acrodef{MSE}[MSE]{Mean Squared Error}
\acrodef{LV}[LV]{Left Ventricle}
\acrodef{mAP}[mAP]{Mean Average Precision}
\acrodef{NF}[NF]{Normalising Flow}

\title{ConceptVAE: Self-Supervised Fine-Grained Concept Disentanglement from 2D Echocardiographies}
\date{}

\author{\hl{Costin F. Ciușdel} 
 $^{1,}$*, Alex Serban $^{1}$ and Tiziano Passerini $^{2}$}

\author{ {\hspace{1mm}Costin F. Ciusdel*, \space\space Alex Serban} \\
	Foundational Technologies\\
	Siemens SRL, Brasov, Romania\\
	\texttt{costin.ciusdel@siemens.com} \\
	\texttt{alexandru.serban@siemens.com} \\
    \And
    {\hspace{1mm}Tiziano Passerini} \\
    Siemens Healthineers\\
    Princeton, NJ, USA\\
    \texttt{tiziano.passerini@siemens-healthineers.com} \\
}

\renewcommand{\headeright}{}
\renewcommand{\undertitle}{}

\begin{document}
\maketitle

\begin{abstract}
While traditional self-supervised learning methods improve performance and robustness across various medical tasks, they rely on single-vector embeddings that may not capture fine-grained concepts such as anatomical structures or organs. 
The ability to identify such concepts and their characteristics without supervision has the potential to improve pre-training methods, and enable novel applications such as fine-grained image retrieval and concept-based outlier detection.
In this paper, we introduce ConceptVAE, a novel pre-training framework that detects and disentangles fine-grained concepts from their style characteristics in a self-supervised manner. 
We present a suite of loss terms and model architecture primitives designed to discretise input data into a preset number of concepts along with their local style.
We validate ConceptVAE both qualitatively and quantitatively, demonstrating its ability to detect fine-grained anatomical structures such as blood pools and septum walls from 2D cardiac echocardiographies. 
Quantitatively, ConceptVAE outperforms traditional self-supervised methods in tasks such as region-based instance retrieval, semantic segmentation, out-of-distribution detection, and object detection.
Additionally, we explore the generation of in-distribution synthetic data that maintains the same concepts as the training data but with distinct styles, highlighting its potential for more calibrated data generation. 
Overall, our study introduces and validates a promising new pre-training technique {based on concept-style disentanglement}, opening multiple avenues for developing {models for medical image analysis that are more interpretable and explainable than black-box approaches}.   
\end{abstract}



\section{Introduction}
\label{sec:intro}

Unsupervised and, in~particular, \ac{SSL} methods facilitate the  use of unlabeled data to learn its underlying structure. 
These pre-training methods have demonstrated improved performance and robustness across a wide range of medical imaging tasks, outperforming models trained solely through supervised learning~\cite{taleb20203d, azizi2021big, huang2024systematic}. 

The core idea of \ac{SSL} pre-training is to develop meaningful representations from input samples, represented as a single continuous embedding vector encapsulating the content displayed in an input~\cite{ssl_cookbook}.
These representations can be viewed as an aggregation of local concepts, their corresponding styles and their contribution on the overall meaning of the input. 
The nature of the representations learnt can vary depending on the specific method employed~\cite{cabannes2023ssl}. 
For example, some methods encourage the representations to be similar for similar or augmented input samples, and~dissimilar for samples that depict distinct concepts~\cite{wu2024voco}.
Other methods aim to ensure that the representations can be accurately reconstructed from partially masked inputs or features~\cite{baevski2022data2vec, wang2023swinmm}.

Regardless of the approach employed, each method aims to develop a single-vector representation of the input, which may fail to capture fine-grained concepts present in it.
For example, a~2D echocardiography of the heart can be broken down into concepts such as heart chambers, valves, and~walls. 
However, the~\ac{SSL} methods' single-vector representation makes it challenging to discern whether such concepts are learned during pre-training~\cite{liu2024benchmarking,holste2024efficient}. 

Moreover, similarity constraints imposed in \ac{SSL} under various augmentations can cause algorithms to merge certain concepts and their associated styles. 
For example, two augmented views of the same input must produce similar representations. 
However, cropping or zooming can exclude some object parts from a view, while blurring or color jittering can alter local textures, making them different between the augmented views. 
This is one reason why \ac{SSL} pre-trained models typically do not perform well on localized tasks, such as detecting localized pathologies, instance retrieval or \ac{OOD} detection~\cite{insclr,retrieval_survey}. 
The ability to identify individual concepts that make up larger objects within input images, and~capture particular traits of these concepts such as textures, will result in more expressive embeddings that can alleviate some of these~weaknesses.

\begin{figure}[b]
    \centering
    \includegraphics[width=8.5cm, keepaspectratio]{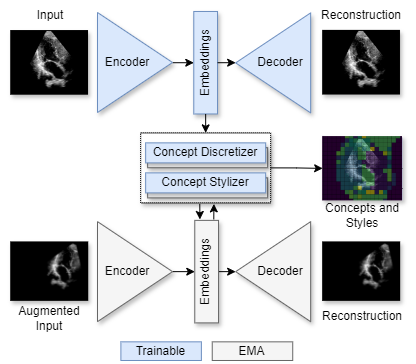}
    \caption{\hl{ConceptVAE} 
 overview, where the blue blocks are trainable while the grey blocks are only updated using exponential moving~average.}
    \label{fig:conceptvae_overview}
\end{figure}

In this paper, we present a novel pre-training method that learns to discretise an input image into a set of fine-grained concepts, and~identifies a unique set of styles for each concept. 
Inspired by human perception, where the brain rapidly recognizes objects by first identifying essential concepts as key components and then perceiving detailed information like fine textures, our approach aims to mimic this process~\cite{human_psych,dicarlo2012does,wardle2020recent}. 
Using 2D cardiac echocardiographies, we show that the proposed method, which we term ConceptVAE and illustrate in Figure~\ref{fig:conceptvae_overview}, can identify fine-grained concepts representing anatomical structures and regions such as heart chambers, walls or blood pools without any~supervision.

{The main strength of our proposed framework is the concept(content)–style disentanglement that happens natively during the pretraining procedure, a~behavior that doesn’t occur within traditional \ac{SSL} methods. We demonstrate the achievement of disentanglement and investigate its potential in a plurality of diverse downstream tasks (such as segmentation, object detection, retrieval, generation, outlier detection) where we directly exploit the proposed disentangled latent space. Applications in medical imaging, where aspects such as model explainability and interpretability hold great interest, can benefit from concept-style disentanglement of the latent space. 
Although traditional \ac{DL} models are capable of performing the aforementioned tasks with good performance, they lack such properties since they are black-box solutions (regardless whether pretraining was used or not in their development). Disentanglement can also be used as a tool to explore the underlying structure of data, through the explicit decomposition into observed local concepts and their style properties.} 

Briefly, ConceptVAE extends the \ac{VAE}  framework to encode a 2D input image into a latent space using a 2D grid of concept probability distributions (one \(p_{ij}(c)\) for each image region, where $c$ is a concept and $i, j$ are spatial indexes) and their associated style vectors (\(s_{ij} = f(c_{ij}, x)\), where $s_{ij}$ is the style property vector of concept $c_{ij}$ that is present at location $i, j$ in input image $x$).
We find that even a modest number of discrete concepts and styles (\eg 16 concepts and 8 style components) are sufficient to model 2D echocardiographies. 
We design a series of loss functions that guide a neural network to detect underlying concepts from an input image and identify particular styles for each~concept.

We validate the effectiveness of the embeddings learnt via ConceptVAE through distinct tasks including region-based instance retrieval, semantic segmentation, object detection, and~\ac{OOD} detection, demonstrating consistent improvements over more traditional \ac{SSL} methods.

In summary, our work's key contributions are the following:

\begin{itemize}
	\item  We introduce ConceptVAE, a~novel \ac{SSL} training framework that yields models capable to fine-grained disentangle concepts and styles from medical images. We evaluate the model using 2D cardiac echocardiographies, given the accessibility of datasets for pre-training and validation. Nevertheless, ConceptVAE is designed to be versatile and can potentially be applied to all 2D image modalities.
    \item We qualitatively validate ConceptVAE and demonstrate its ability to identify concepts specialised for anatomical structures, such as blood pools or septum walls.
    \item  We quantitatively validate ConceptVAE and show consistent improvements over traditional \ac{SSL} methods across various tasks, including instance retrieval, semantic segmentation, object detection, and~\ac{OOD} detection.
    \item We assess ConceptVAE's ability to generate data conditioned on concept semantics and discuss its potential to enhance robustness in dense prediction tasks.
\end{itemize}

The remainder of this article is organised as follows.
We start by discussing background information and related work (Section~\ref{sec:related_work}), followed by a detailed overview of ConceptVAE (Section~\ref{sec:concept_vae}), an analysis of the pre-trained model's ability to disentangle concepts and styles (Section~\ref{sec_model_analysis}), and~a quantitative evaluation of the model for multiple tasks (Section~\ref{qualitative_analysis}).
The paper ends with conclusions and future work (Section~\ref{sec:conclusions}).

\section{Related~Work}
\label{sec:related_work}

We identify a series of related works that can be categorized into three distinct groups: (i) \ac{SSL} methods, encompassing both general approaches from natural images and those specific to medical images~\cite{ssl_cookbook,zhang2023dive}; (ii) \ac{DRL} methods, which aim to train models capable of identifying and mapping factors of variation to semantically meaningful variables~\cite{eddahmani2023unsupervised,disentangled_1}; and (iii) the application of \ac{SSL} methods to improve performance in medical image processing tasks related to 2D echocardiographies, such as segmentation or information retrieval. 
Below, we discuss these groups independently and explore their~interplay.

The primary \ac{SSL} methods can be categorized in (i) contrastive learning methods \mbox{(\eg~\cite{simclr,moco}}), which aim to create similar representations for input images showing the same objects and contrastive representations for images showing different objects; (ii)~correlation-based methods (\eg~\cite{vicreg}), which aim to preserve the variance of the embeddings while decorrelating variables related to distinct objects; and (iii) masked image modeling methods (\eg~\cite{masked_autoencoder,data2vec}), which aim to reconstruct the original input from its masked version.
Recent studies indicate that, despite differences in methodology and training objectives, contrastive and correlation-based methods are closely related and may yield similar results, as~they minimize criteria that are equivalent under certain conditions~\cite{self_duality}.
All methods in these groups focus on developing single-vector (and not local or concept-based) representations, which can be used in distinct downstream~tasks.

Within \ac{SSL} methods, some approaches yield models with interesting emergent properties. 
For example, vision transformer models~\cite{vit} trained with DINO~\cite{emerging_props, tian2024learning} can generate features that explicitly describe the semantic segmentation of an image. 
These features can be directly linked to actual objects present in the image, which can be broadly interpreted as independent concepts. 
Training with DINO improves performance in image classification, segmentation, and~even information retrieval. 
Building upon DINO, ref.~\cite{acseg} associated a fixed number of prototypical concepts with the semantics of each image using a pixel assignment scheme based on k-means clustering, further enhancing semantic~segmentation.

Despite the fact that global representations developed through \ac{SSL} methods can linearly separate certain object classes, these methods do not ensure that the learned latent space structure is meaningful. 
Specifically, intermediate feature maps (\ie~the spatial feature maps before the final projector head) may not be sufficiently descriptive to reliably differentiate between similar visual concepts or to group together representations of objects from the same class.
Additionally, these representations might either be intertwined with style information or attempt to suppress it to achieve invariance against train-time augmentations~\cite{disentangled_1}. 

In contrast, \ac{DRL}  is a family of training methods aimed at isolating the factors of variation driving the generative process behind a data distribution into distinct latent variables.
Ref.~\cite{disentangled_1,disentangled_2} provide overviews of recent techniques in \ac{DRL}. 
Among various benefits, \ac{DRL} can improve a model's explainability, controllability, and~robustness~\cite{disentangled_2}.
Nevertheless, \ac{DRL} methods often need labels to learn meaningful representations~\cite{locatello2019challenging} and have limited applicability to image-based tasks, primarily focusing on image generation~\cite{disentangled_2}. 
In contrast, ConceptVAE is designed as a general pre-training strategy that benefits multiple downstream~tasks.

Within \ac{DRL}, ConceptVAE is similar to \emph{\hl{content-style disentanglement}}~\cite{disentangled_1},   
 as~it deliberately assigns distinct roles to different components of the latent space. 
For example, certain components represent anatomical concepts such as heart valves (acting as the \emph{content}), while others capture their local specifics (acting as the \emph{style}).
Our model uses both discrete and continuous latent variables, for~the content and style of input images, respectively. 
This approach has proven successful in other \ac{DRL} works, \eg~for clustering latent space representations in generative adversarial modeling~\cite{cluster_gan}. 
However, our two latent variables are not independent: the style is determined as a function of both the input image and a predicted grid of discrete~concepts.

While some methods enforce DRL at train time through either inductive biases, priors or supervision~\cite{disentangled_1}, other methods work post-hoc as post-processing of pretrained models in order to separate style and content. 
For example, ref.~\cite{style_content_disentang} uses style annotations to compute a linear projection that is applied on the entangled representations to separate them in two sub-matrices: a diagonal style matrix and an invertible dense content matrix.
We draw inspiration from this approach, and~enforce a unit-covariance constraint on the style component of our latent space, while letting adjacent concepts cooperate for reconstructing the input~image.

Modeling images with a discrete codebook has been previously employed for purely generative purposes in models such as VQ-VAE~\cite{vqvae2,taming}. 
Unlike our approach, these models require a significantly larger codebook size because a discrete code must represent a combination of entangled concept and style. 
In contrast, our model requires only a small array of discrete concepts, as~they are disentangled from the styles, which are represented in the latent space by small-sized continuous~vectors.

Similar methods have been employed in cardiac image analysis before. 
For example, ref.~\cite{chartsias}~used spatial binary anatomical factors as content to compute an image-level modality factor as style for reconstructing MRI and CT data. 
Additionally, traditional \ac{SSL} methods have been successfully applied in medical image analysis for tasks such as instance retrieval~\cite{wang2023retccl}, semantic segmentation~\cite{fischer2023self}, and~object detection~\cite{zhang2023dive}. 
However, these models are adapted from natural image analysis and are not specifically tailored for medical~imaging.



\section{ConceptVAE}
\label{sec:concept_vae}

Figure~\ref{fig:conceptvae_overview} presents a high-level overview  of ConceptVAE. 
In essence, the~method employs a \ac{VAE}-like architecture to reconstruct an input from the model's embeddings. 
It then converts the features into a set of concepts and styles via the concept discretizer and concept stylizer~blocks.

We include a self-supervised input reconstruction task because we train the model from scratch and require an encoder that can produce meaningful low-level embeddings. 
However, this task is separated (through a stop-gradient operation) from  concept and style identification.
Using an existing pre-trained encoder can replace this~task.

To prevent feature collapse, such as unique features for all inputs or a single concept for all concept maps, as~well as improve training stability, we use a mirrored network for augmented versions of the input, updating it only with \ac{EMA}---a technique proven in \ac{SSL} methods with similar aims~\cite{emerging_props}.

Both the original and augmented input embeddings are transformed, discretized and styled using the concept discretizer and stylizer blocks. 
To ensure consistency in concepts between augmented versions of the input, a~specialized loss term is employed.
To guide the model in learning significant concepts and styles, the~original inputs are reconstructed from the concepts and styles using the \ac{EMA} decoder. 
A dedicated reconstruction loss term is employed to ensure that the inputs reconstructed from concepts and styles closely match the originals. 
This process encourages the model to capture and represent meaningful features of the data within the learned concepts and styles.
Similarly, localised loss terms guide the model to learn diverse concepts and~styles.

The following subsections elaborate on the architecture, the~rationale behind its design, and~the training procedure, including details about the  selected loss function terms and optimization~parameters.

\subsection{Model~Architecture}
\label{subsec:concept_vae}

\begin{figure}[t]
	
	\includegraphics[width=\linewidth, keepaspectratio]{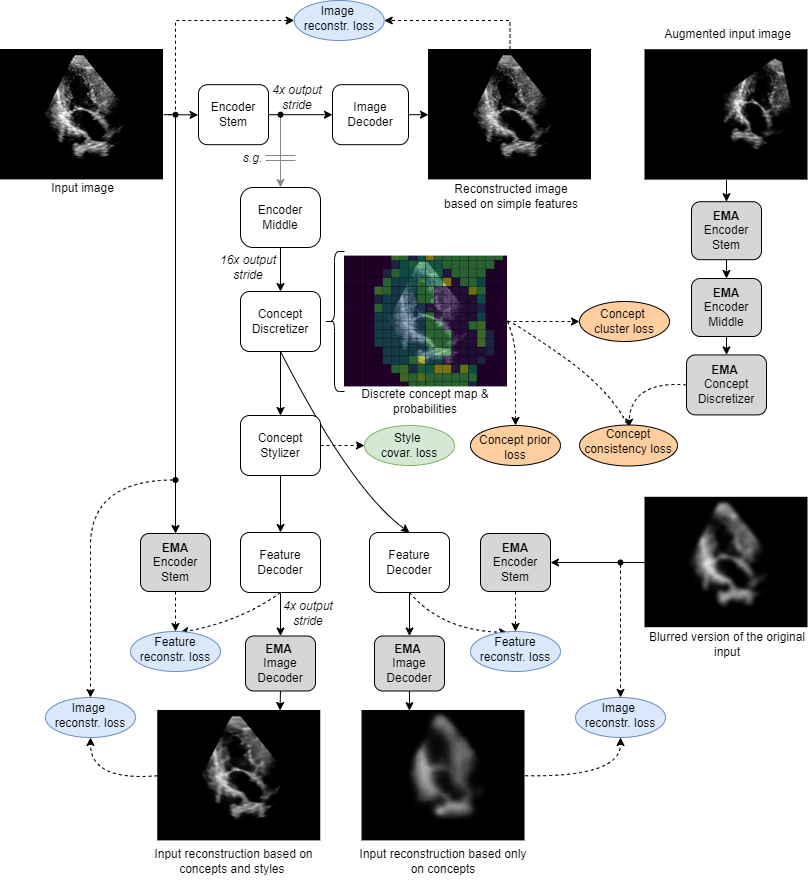}
	
	\caption{ ConceptVAE model architecture and training setup, where the EMA blocks represent the exponential moving average mirrors of regular blocks. Loss components are shown in colored ellipses, and~\textit{s.g.} denotes \textit{stop-gradient}. Solid arrows indicate tensor flows within the model, while dashed arrows represent tensors involved in loss~functions.}
	\label{fig1}
\end{figure}

Figure~\ref{fig1} displays the detailed architecture of ConceptVAE. 
A simple auto-encoder operates independently (in terms of gradients) from the rest of the model. 
It comprises an \emph{Encoder Stem} that generates features $x_{stem}$ at a 4$\times$ 
 output stride, and~an \emph{Image Decoder} that reconstructs the original input. 
After a stop-gradient operation, an~\emph{Encoder Middle} block applies a series of residual convolutional blocks starting from the encoder stem's features, projecting the features to~concepts.

The projections are used by a \emph{Concept Discretizer} classification head, with~$x_{middle}$ having a 16$\times$ output stride. 
For each spatial location, a~Softmax activation creates a probability distribution over 
$C$ concepts. 
Using the Gumbel-Softmax trick~\cite{gumbel} with hard sampling and gradient pass-through, a~grid of one-hot vectors is sampled from the concept probabilities grid. 
This one-hot vector grid indexes a learned matrix of concept embeddings to produce a 2D concept map $x_{concept}$.

Subsequently, $x_{middle}$ and $x_{concept}$ are concatenated along the channel axis and passed into a \emph{Concept Stylizer} block.
This block generates a 2D grid $x_{style}$ of $S$ channels capturing the style properties of each concept. 
At this point, each location within the 16$\times$-stride grid has an identified concept and an associated style vector. 
The channel-wise concatenation of $x_{concept}$ and $x_{style}$ constitutes the model's latent space ($x_{latent}$). 
Notably, $x_{concept}$ is derived from discrete embeddings, using a shared learnable  embedding matrix for all input samples.
In contrast, $x_{style}$ is a continuous tensor computed based on local features  $x_{middle}$ and the sampled discrete concepts $x_{concept}$. 
Consequently, $x_{style}$ is specific to the sampled $x_{concept}$, meaning that sampling a different concept at location $i, j$ will result in a different style vector $x_{style}^{ij}$.

A \emph{Feature Decoder} projects the latent space to reconstruct the lower 4$\times$-stride features of the \emph{Encoder Stem}, denoted as $x_{stem}^{rec}$.
Lastly, the~\emph{\ac{EMA} Decoder} is employed to recover the original input image from the latent space. 
This reconstruction is core to ConceptVAE, as~it guides the model to learn how to decompose an input into fine-grained concepts with associated styles, and~reconstruct the input from concepts alone or from concepts and associated styles.
Using the \emph{\ac{EMA} Decoder} for the reconstruction ensures there is no mode collapse for the concepts or~styles.

Architecturally, the~\emph{Encoder Stem} module is designed as a simple sequence of convolutional, instance normalization, max-pooling, and~Leaky ReLU stages. 
The final layer is a normalization layer that ensures channel-wise zero mean and unit standard deviation, helping to prevent potential feature collapse. 
This module contains three convolutional layers with $3\times3$ kernels and strides \hl{2, 1, 1} 
 respectively, and~one max-pooling layer with $2\times2$ kernel and stride 2, yielding a field of view size of 17 px. 
The \emph{Image Decoder} block maintains this simplicity, consisting of 2 upsampling stages based on $3\times3$ transposed convolution layers with stride 2. 
Regular $1\times1$ convolutions, normalization, and~Leaky ReLU layers are inter-twined between the two up-sampling stages to improve the module's decoding~capacity.

The \emph{Encoder Middle} block employs a residual architecture. 
As in the \emph{Image Decoder} block, the~first layer is a Leaky ReLU activation, as~the input to this block comes from the normalized convolutional output of the Encoder Stem.
The block comprises three residual stages with 3, 5, and~5 residual layers, respectively. Each residual layer includes two sequences of normalization, Leaky ReLU, and~convolution. 
Max-pooling and normalization layers are positioned between each residual stage. 
This  number of layers was selected to ensure that the receptive field-of-view $x_{middle}$
exceeds the shorter dimension of the input image. 
In our case, the~input image has dimensions (h, w) = (256, 320), and~the field of view is approximately 300 pixels.
Larger or smaller architectures can be selected to model distinct input~dimensions.

Equation~(\ref{eq_1}) describes the operation of the concept discretizer. 
A classification head $f_{cd}$ computes the concept probability logits; Gumbel noise $-\ln(-\ln(u))$ is added, and~a temperature ($T_{samp}$) Softmax computes the sampled concept ratios.
A one-hot vector is created based on the concept with largest ratio and the pass-through technique ensures differentiability (where $\sg$ is the stop-gradient operator, $\mathcal{I}$ is the input \hl{image).} 
\begin{equation}
	\begin{aligned}
		p(c)|\mathcal{I} &= Softmax(f_{cd}(x_{middle}(\mathcal{I}))) \\
		u &\sim U(0, 1) \\
		p_{samp}(c) &= Softmax\left(\frac{\ln(p(c)|\mathcal{I}) -\ln(-\ln(u))}{T_{samp}}\right) \\
		y_{hard} &= \mathbf{1}_{hot}(\argmax(p_{samp}(c))) \\
		y_{hard} &= \sg(y_{hard} - p_{samp}(c)) + p_{samp}(c)
		\label{eq_1}
	\end{aligned}
\end{equation}

The \emph{Concept Stylizer} is based on a small 3-layer sequence of convolution---Leaky ReLU---convolution layers, all with bottleneck ($1\times1$) \hl{kernels.} 
Its function is to customize the selected concept at each spatial location within the 16$\times$-stride~grid. 

The \emph{Feature Decoder} begins with two residual stages that process $x_{latent}$, followed by two transposed convolution stages that up-sample the grid to a 4$\times$ output stride relative to the input size. 
These two residual stages operate on a neighborhood of 5 $\times$ 5 spatial locations, allowing adjacent concepts to collaborate in the reconstruction. 
The impact of neighborhood size on reconstruction and modeling quality is discussed in Section~\ref{sec_model_analysis}.

Neither the \emph{Image} nor the \emph{Feature Decoder} employ skip-connections that reuse internal encoder feature maps. 
This design is essential, as~it compels the model to rely solely on its latent space, $x_{latent}$, to~represent the data manifold and reconstruct the inputs.

\subsection{Training~Objectives}
\label{subsec:training_section}

To train ConceptVAE, we devise a series of loss terms inspired by classical (discrete) \ac{VAE} formulations, but~adapted to guide the learning process towards identifying and personalizing concepts. 
We employ two types of reconstruction losses, illustrated in blue in Figure~\ref{fig1}: an image-based loss  $\mathcal{L}_{img}$, which uses \ac{MSE} over pixel values, and~a feature-based loss $\mathcal{L}_{feat}$, which uses \ac{MSE} over low-level feature tensors.
The simple auto-encoder is trained using $\mathcal{L}_{img}$ between the original input image  $\mathcal{I}_{orig}$ and the reconstructed image based on the 4$\times$-stride feature map. 
The \ac{EMA} version of the \emph{Encoder Stem} is used to compute the target for the tensor produced by the \emph{Feature Decoder} block, while the \emph{\ac{EMA} Decoder} is used to compute the reconstructed image  from $x_{latent}$.
The use of both pixel- and feature-level reconstruction losses has been previously employed in VAE/GAN setups~\cite{taming,spade}, {to boost both training stability and image generation fidelity}.

The feature decoder takes both $x_{concept}$ and $x_{style}$ as inputs. 
While $x_{concept}$ is generated by sampling from a discrete concept codebook, $x_{style}$ is computed directly as a (continuous) function of $x_{middle}$ and $x_{concept}$.
Consequently, the~network could potentially exploit this setup by minimizing the influence of $x_{concept}$ and  relying more heavily on the more direct path of $x_{style}$, effectively reducing its operation to that of a simple auto-encoder.
In this scenario, $x_{concept}$ would lose its semantic significance, and~$x_{style}$ would function as a rich bottleneck representation rather than a style characteristic of a concept. 
To address this undesired behavior, an~image/feature reconstruction is performed where the style components of $x_{latent}$ are explicitly zeroed out. 
The \emph{\ac{EMA} Decoder} is reused to obtain a reconstructed version of the input image, relying solely on $x_{concept}$, without~the style component $x_{style}$. 
The target of this reconstruction is a blurred version of the input image, with~blurring serving as an approximation for removing fine details and textures, thereby partially eliminating the notion of style. 
Both pixel- and feature-based losses are employed to evaluate the reconstruction quality when using only the spatial distribution of concepts. 
This approach guides the \emph{Feature Decoder} block to focus on the concept component of $x_{latent}$ and also encourages the \emph{Encoder Middle} to learn to detect relevant concepts within input~images.

Another key aspect of concept detection is its invariance to specific styles. 
This means that two different (augmented) views of the same medical image should produce the same concept maps, despite variations in their  visual appearances. 
Pixel-level and texture differences should be captured by $x_{style}$, while more complex anatomical structures should be encoded in $x_{concept}$. 
To guide this behavior during training, we introduced a \emph{Concept consistency loss}, illustrated with orange in Figure~\ref{fig1}.
The \emph{Concept Discretizer} block first computes a grid of concept probabilities, from~which it generates a spatial grid of sampled concept indices.
Following this, the~concept maps from augmented views should be equivalent, even if the augmentations involve translations, rotations, or~other spatial shifts ({\hl{We} 
 use \emph{equivalent} instead of \emph{identical} because augmentations like translations, rotations, and~shearing can spatially shift the placement of concepts within the image. Nevertheless, the~correspondences between the initial and shifted locations are known, and~they can be used to enforce similarity between $p(c) | \mathcal{I}_{orig}$ and $p(c) | \mathcal{I}_{augm}$}).

The \emph{\ac{EMA} Encoder Stem}, \emph{\ac{EMA} Encoder Middle}, and~the \emph{\ac{EMA} Concept Discretizer} are used to compute the target probability distributions $p_{ema}(c)$ for the concept consistency loss: $\mathcal{L}_{cc} = - p_{ema}(c)\ln p(c)$. 
The \ac{EMA} concept probability map $p_{ema}(c)$ is computed on an augmented view of the initial input image which incorporates transformations such as rotations, translations, shearings, zooming, gamma contrast changing and Gaussian blurring. 
Since these operations can alter positions, we must account for the spatial  mapping between $p(c)$ and $p_{ema}(c)$. 
To simplify this  and avoid optimization noise due to imperfect mapping,  each augmentation procedure selects a random location uniformly, and~all image operations are performed relative to this point. 
The result includes a tuple of the augmented input image $\mathcal{I}_{augm}$, an~initial location $l_{ij}$, and~the equivalent location $l_{i'j'}$ after all operations. 
In our implementation of $\mathcal{L}_{cc}$ we indexed only the grid positions of the spatial locations $l_{ij}$ and $l_{i'j'}$ from $p(c)$ and $p_{ema}(c)$, respectively. 
Therefore, only one pair of grid locations (containing the concept probability distributions) is used per each sample inside a training batch. 
We use the \ac{EMA} blocks instead of the model blocks to prevent feedback loops that could lead to collapsing concept probabilities (\eg~always detecting the same concept). 

An additional constraint $\mathcal{L}_{style}$ was imposed on $x_{style}$ to ensure it has  unit covariance and zero mean along the channel (style) dimension (illustrated with green in Figure~\ref{fig1}).
Specifically, when $x_{style}$ is flattened across across batches ($B$), height ($H$) and width ($W$) if forms a matrix of shape shape $(S, BHW)$.
This matrix must have a row-wise mean of 0, a~row-wise standard deviation of 1, and~zero correlation between rows. 
This constraint ensures that $x_{style}$ has independent components with a known range of values, discussed in details in Section~\ref{subsec_style_augm}. 

To control the deviation of $p(c)|\mathcal{I}$ from $p_0(c)$, we use two priors. 
Without enforcing these priors during training, the~entropy of $p_{ij}(c)$ would be minimized, canceling the effect of concept sampling and reducing the model's operation to a deterministic auto-encoder.  
Consequently, the~concept probability grid  $p(c)|\mathcal{I}$ would lose much of its semantic significance, reverting to a regular discrete latent variable instead of encoding high-level semantics into a fixed set of concept probabilities. 
This, in~turn, would constrain the functionality of the concept consistency loss.
We employ two types of priors: at the grid-location level and at image level. 
Since we are modeling echocardiographies, these images typically feature an ultrasound cone centered within a surrounding black background. 
The grid-location level prior  is computed as follows: for grid locations inside the ultrasound cone, the~prior is a uniform distribution over the last $C-1$ concepts, with~ the first concept having zero mass (as we always designate the first concept to model the background).
For grid locations outside the cone, the~prior assigns all probability mass to the first~concept. 

The KL-divergence $\mathcal{D}_{KL}\infdivx{p(c)|\mathcal{I}}{p_0(c)}$ is computed at all grid locations and averaged across the $(B, H, W)$ dimensions. 
For the image-level prior loss it is assumed that only the first concept should be detected outside the cone, with~a uniform spread of concepts inside the cone across all samples in the current batch. 
Therefore, the~concept probability vectors of all grid locations inside and outside the echo cones are averaged across all samples in the batch to obtain two image-level concept prevalence vectors: $d_{cone}(c)$ for the cone region and $d_{bg}(c)$ for the~background. 

The KL-divergence loss with the same priors is used for these concept prevalence vectors. 
Equation~(\ref{eq_prior}) formalizes the final prior loss $\mathcal{L}_{prior}$, where $\mathbf{1}_{c}(b,i,j)$ is an indicator function that equals 1 if location $i, j$ in sample $b$ of the current batch pertains to an ultrasound cone. 
$N_{cone}$ and $N_{bg}$ are the total numbers of cone and background grid locations inside current batch, respectively.

\begin{align}\label{eq_prior}
	\begin{split}\nonumber
		\mathcal{L}_{prior1} ={}& \sum_{b,i,j}  \frac{\alpha_1}{N_{cone}}\mathcal{D}_{KL}\infdivx{p_{bij}(c)|\mathcal{I}}{p_0^{cone}(c)} \mathbf{1}_{c}(b,i,j) \\
		& + \frac{\alpha_2}{N_{bg}}\mathcal{D}_{KL}\infdivx{p_{bij}(c)|\mathcal{I}}{p_0^{bg}(c)}  (1-\mathbf{1}_{c}(b,i,j))
	\end{split}\\
	\begin{split}\nonumber
		d_{cone}(c) ={}& \frac{1}{N_{cone}} \sum_{b,i,j} (p_{bij}(c)|\mathcal{I})\mathbf{1}_{c}(b,i,j)
	\end{split}\\
	\begin{split}\nonumber
		d_{bg}(c) ={}& \frac{1}{N_{bg}} \sum_{b,i,j} (p_{bij}(c)|\mathcal{I})(1-\mathbf{1}_{c}(b,i,j))
	\end{split}\\
	\begin{split} \nonumber
		\mathcal{L}_{prior2} ={}& \alpha_3		\mathcal{D}_{KL}\infdivx{d_{cone}(c)}{p_0^{cone}(c)} \\
		& + \alpha_4\mathcal{D}_{KL}\infdivx{d_{bg}(c)}{p_0^{bg}(c)}
	\end{split}\\
	\begin{split}
		\mathcal{L}_{prior} ={}& \mathcal{L}_{prior1}+\mathcal{L}_{prior2}
	\end{split}
\end{align}

To discourage overly granular concept maps, where sampled concepts change frequently between adjacent grid location, we use a \emph{Concept cluster loss} $\mathcal{L}_{cluster}$, depicted in orange in Figure~\ref{fig1}).
Overly granular concepts are undesirable because we want concepts to represent larger anatomical structures spanning multiple grid locations rather than smaller, granular pixel patterns. 
To enforce it, we use the one-hot vectors produced by the \emph{Concept Discretizer} block.
We compute spatial derivatives between adjacent one-hot vectors along the width and height dimensions.
If two adjacent locations share the same sampled concept their one-hot vectors are identical, resulting in a null spatial derivative.
Otherwise, the~sampled concepts differ, leading to different one-hot vectors and a nonzero spatial derivative.
By minimizing the mean square of the spatial derivative, we reduce the number of spatial transitions between sampled concepts, thereby creating larger concept “islands”. 
The mean is taken only over grid-locations pertaining to ultrasound~cones. 

The final loss function is a weighted sum of the described sub-losses, as~shown in Equation~(\ref{eq_full_loss}).
Here, $f_{dec}(x)$ denotes the feature computed by the \emph{Feature Decoder} block based on its input $x$, and~$\mathcal{I}_{rec}([x_{concept}, x_{style}])$ represents the reconstructed image based on latent space components $x_{concept}$ and $x_{style}$. 
\begin{equation}\label{eq_full_loss}
\begin{aligned}
	\mathcal{L} = &\beta_1\mathcal{L}_{img}(\mathcal{I}_{rec}(x_{stem}), \mathcal{I}) +\\
	&\beta_2\mathcal{L}_{img}(\mathcal{I}_{rec}([x_{concept}, x_{style}]), \mathcal{I}) +\\
	&\beta_3\mathcal{L}_{img}(\mathcal{I}_{rec}([x_{concept}, x_{style}:=0]), \mathcal{I}_{blurred}) +\\
	&\beta_4\mathcal{L}_{feat}(f_{dec}([x_{concept}, x_{style}]), f_{stem}(\mathcal{I})) +\\
	&\beta_5\mathcal{L}_{feat}(f_{dec}([x_{concept}, x_{style}:=0]), f_{stem}(\mathcal{I}_{blurred})) +\\
	&\beta_6\mathcal{L}_{style}(x_{style})  +\\
	&\beta_7\mathcal{L}_{cc}(p(c)|\mathcal{I}, p_{ema}(c)|\mathcal{I}_{augm}) +\\
	&\beta_8\mathcal{L}_{prior}(p(c)|\mathcal{I}) +\\
	&\beta_9\mathcal{L}_{cluster}(x_{concept})
\end{aligned}
\end{equation}

\subsection{Pre-Training Data and~Hyper-Parameters}
\label{subsec:pretraining}

To pre-train ConceptVAE, we used 72,500 frames extracted from 7500 echocardiography video acquisitions. 
The dataset consisted exclusively of 2D B-mode echocardiographies featuring apical or short-axis~views.

We used the AdamW optimizer with a constant learning rate of 
$10^{-4}$, a~batch size of 64 images, and~a weight decay of 
\hl{$5 \times 10^{-3}$.} 
During training, we apply random image augmentations using the following transformations: rotation, translation, shearing, zooming, gamma contrast adjustment, and~Gaussian blurring. 
Pre-training is performed until convergence, which is equivalent to the loss function no longer varying~significantly.

\section {Latent Space and Qualitative Analysis}
\label{sec_model_analysis}

\begin{figure}[b]
	 
	\begin{subfigure}{0.325\linewidth}
		\includegraphics[width=\linewidth]{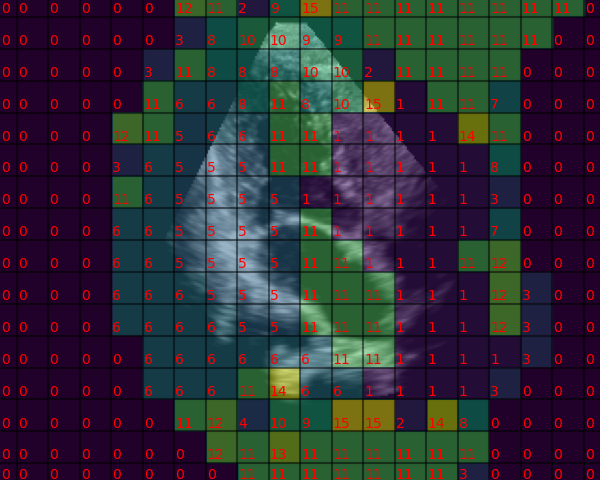}
	\end{subfigure}
	\begin{subfigure}{0.325\linewidth}
		\includegraphics[width=\linewidth]{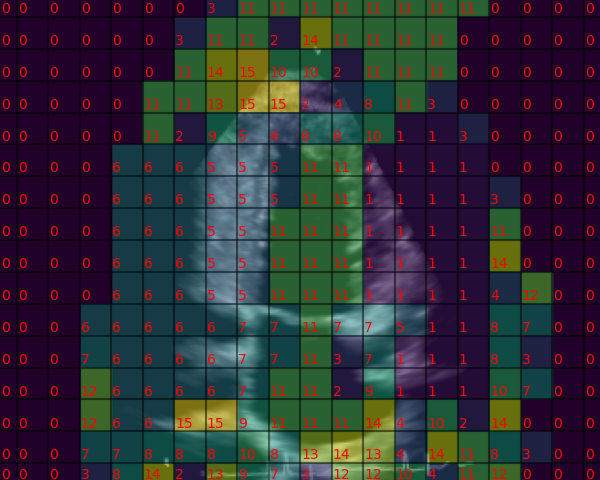}
	\end{subfigure}
	\begin{subfigure}{0.325\linewidth}
		\includegraphics[width=\linewidth]{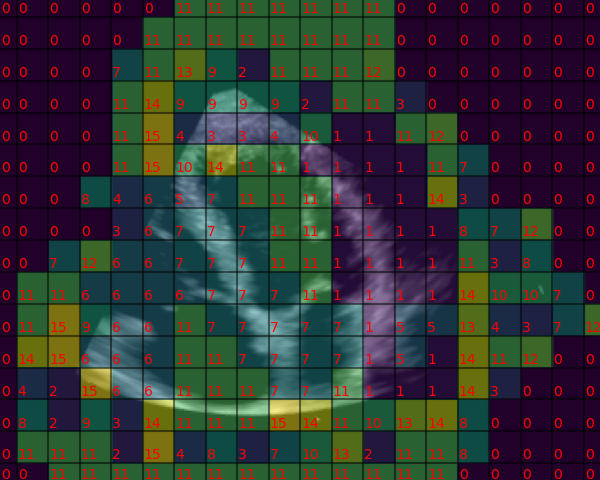}
	\end{subfigure}
	\caption{\hl{Concept} 
 maps for three randomly sampled inputs. The~16$\times$-stride concept grid is up-sampled to the original image size. The~indices of the most likely concept for each grid location are displayed in red at the bottom-left of each location. The~grid is color-coded according to concept indices for better~visualisation.}
	\label{fig_concept}
\end{figure}

Upon convergence, the~pre-trained model can be qualitatively analysed by examining the inferred concept probability maps for test images. 
A straightforward method to implement this involves selecting the most likely concept at each grid location \mbox{($c_{ij} = \argmax p_{ij}(c)$}) and overlaying the up-sampled concept indices grid onto the initial input images, as~in Figure~\ref{fig_concept}. 
The probability of the most likely concept $p(c_{ij}) = \max p(c)$ at each location $i, j$ can be incorporated in the~visualisations. 

By examining a random selection of samples illustrated in Figure~\ref{fig_concept}, we can make the following initial observations:

\begin{itemize}
	\item The prior constraint, which requires regions outside the cone to be modeled solely by the first concept (\ie~the background concept at index 0) is generally respected. Exceptions occur at  grid locations in the cone's proximity, particularly at the boundaries between the cone and the background. As~these are transition regions, they are not particularly concerning, since the model's confidence is expected to be low for such~regions.

	\item Certain concepts are specialized for specific anatomical structures. For~example,~concept $c_{11}$ models blood pools within the cone, concept $c_{1}$ represents the \ac{LV} free wall on the right hand size of the cone, concepts $c_5$ and $c_7$ correspond to septum walls, and~concept $c_6$ covers the right-heart side of the cone, among~others.
 
	\item Certain concepts, such as \eg~$c_{13}$ and $c_{14}$  appear more isolated and spanning a single grid location. By~qualitatively assessing multiple input samples, we hypothesise these concepts encode information about the local anatomical shapes of nearby larger concept islands. It appears these concepts have larger confidence assigned to them than the average confidence inside larger concept islands. We term them \emph{modifier}~\mbox{concepts}.
    
\end{itemize}

\begin{figure}
 	\centering
 	\begin{subfigure}{0.325\linewidth}
 		\includegraphics[width=\linewidth]{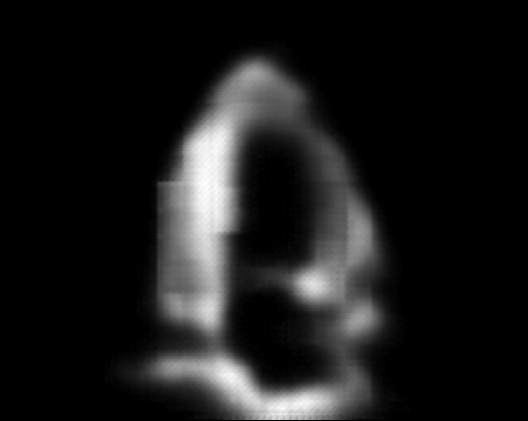}
 	\end{subfigure}
 	\begin{subfigure}{0.325\linewidth}
 		\includegraphics[width=\linewidth]{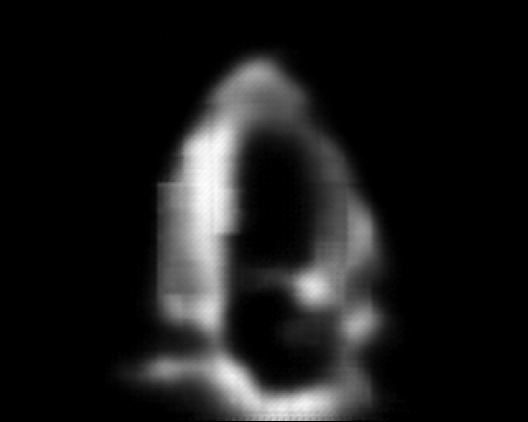}
 	\end{subfigure}
 	\begin{subfigure}{0.325\linewidth}
 		\includegraphics[width=\linewidth]{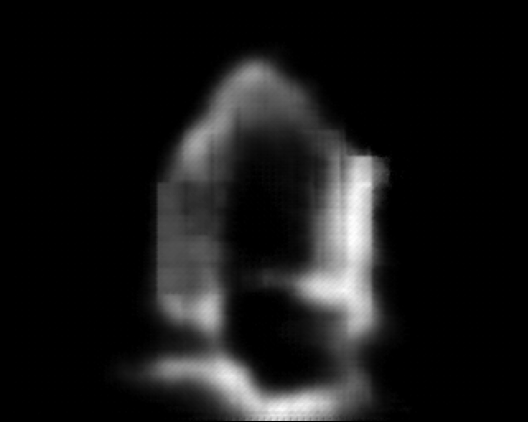}
 	\end{subfigure}
 	\caption{ Effect of concept swapping. The~left image is the reconstruction based only on the greedy concept map (with $x_{style}:=0$). The~middle reconstruction illustrates the effect of swapping 2 modifier concepts, while the right reconstruction illustrates big changes induced by swapping two anatomy-specific~concepts.}
 	\label{cmap_swap}
\end{figure}

To qualitatively evaluate the impact of modifier concepts, the~greedy concept map of the middle image of Figure~\ref{fig_concept} is modified in two ways, by~swapping 2 modifier and 2~normal concepts: first, (i) the modifier concepts $c_{13}$ and $c_{14}$ are swapped and the image is reconstructed without any style component ($x_{style}:=0$); and (ii) starting  from the greedy map, concepts $c_5$ and $c_1$ are now swapped and the image is reconstructed in the same manner (with  $x_{style}:=0$). 
The effects are illustrated in Figure~\ref{cmap_swap}: in the former case only minor shape modifications are observed around the grid locations where concept swaps were done. 
In the latter case, the~effect is more significant, as~it appears that the \ac{LV} free wall changed place with the~septum. 

While modifier concepts seem to function primarily in a styling role, it is important to note that the  \emph{Feature Decoder} block processes $k \times k$ regions of adjacent concept locations to reconstruct the low-level image features $x_{stem}$.
This means that neighboring concepts cooperate to form larger and more complex anatomical structures. 
Modifier concepts are not devoid of semantic meaning, as~our experiments showed that replacing a specialized anatomical concept like $c_1$ with a modifier concept still yields similar reconstructions, albeit with slight alterations in shape and/or region brightness patterns.
Additionally, although~reconstructing images based solely on $x_{concept}$ may produce rough outlines of echocardiographies, suggesting that concepts only encode basic brightness blobs, we later show that the concept probability grid contains rich semantics that can be used in tasks such as instance retrieval (Section~\ref{ss_inst_retrieval}). 

The region size $k$ influences the operation and semantics of concepts. 
In the extreme case of $k=1$, there is no concept cooperation and to match $\mathcal{I}_{rec}([x_{concept}, x_{style}:=0])$ with $\mathcal{I}_{blurred}$, concepts may be incentivised to encode blurred pixel patterns instead of semantic content. 
At the other extreme, where $k$ equals the grid size, each grid location has a full receptive field of view, meaning it can observe the concepts from all other grid locations, regardless of distances (similar to a self-attention layer~\cite{vit}). 
This can be undesirable because the model may rely on non-local relations between concept placements instead of embedding semantic content within each concept.
It would also hinder the extraction of local region descriptors, making it impossible to describe the content of an image crop without retaining the entire concept grid. 
Consequently, tasks such as region-based instance retrieval would be challenging, as~it would not be clear how to construct descriptors focused on specific image~regions. 

We employed $k=5$, meaning the receptive field of view before the up-sampling layers inside the \emph{Feature Decoder} block is 5 $\times$ 5 grid locations of  $x_{latent}$).
The rationale is that $k$ should be large enough to allow $\mathcal{I}_{rec}([x_{concept}, x_{style}:=0])$ to have smooth pixel-level transitions between adjacent concepts and thus be close to $\mathcal{I}_{blurred}$, but~small enough to enable the construction of granular region descriptors and prevent the model from exploiting non-local~\mbox{relations}.

\section{ Quantitative Model~Analysis}
\label{qualitative_analysis}

To assess the representation power of the  model's latent space, its suitability as a general pre-training method, and~the extent of content-style disentanglement, we employ a linear evaluation protocol tailored to \ac{SSL}~(\eg~\cite{simclr,moco,haghighi2024self}) on several distinct~tasks.

For comparison, we used a baseline model trained with Vicreg~\cite{vicreg}, featuring a ResNet50 encoder and a lightweight RefineNet decoder~\cite{RefineNet} for dense tasks. 
This model was pre-trained using the same dataset and configuration (\eg~image sizes) as ConceptVAE (Section~\ref{subsec:pretraining}). 
For all following evaluation tasks, we used the output of the second to last ResNet stage as the baseline latent space (as it has the same output stride as our proposed~\mbox{model}).

The linear evaluation protocol involved freezing the backbone and training only a linear layer on top of the frozen embeddings for specific tasks ranging from object detection to semantic segmentation or \ac{OOD} detection, as~detailed in the following~sections.

\subsection{Region-Based Instance~Retrieval}
\label{ss_inst_retrieval}

Region-based instance retrieval involves searching a database of images for similar samples using only localized descriptors, such as pathologies or anomalies. 
These methods can aid in clinical diagnosis, medical research, trainee education, and~support other tasks by quickly identifying patients with similar anomalies, even when a diagnosis is not yet established~\cite{wang2023retccl,kobayashi2024sketch}. 
\ac{SSL} methods are the most prevalent and effective, using the embeddings of a pre-trained model to cluster images and retrieve those most similar to a query image using nearest neighbors search~\cite{li2018large}.

To use ConceptVAE for this task, we generate image region descriptors by concatenating the 5 $\times$ 5 concept probability vectors from a 5 $\times$ 5 sub-grid centered around a selected query point.
The sub-grid provides context for the query~point.

Using an input image of size (256, 320), the~concept grid has an output stride of 16, resulting in a size of (16, 20) concepts. 
From each test image, we extract an array of (14, 18) key points (i.e., all points with a complete 5 $\times$ 5 neighborhood). 
Since the model was trained with 16 concepts and the descriptor uses a 5 $\times$ 5 grid, each descriptor is a vector of size 400.
For the baseline model, a~similar searching mechanism was used, but~the region descriptor was the feature vector of a 1 $\times$ 1 feature map grid location. 
A single grid location is sufficient for this model, since its feature representation is computed in a continuous manner, without~discrete variables, with~a sufficiently large field of~view.

\begin{figure}
	 
	\includegraphics[width=0.99\linewidth]{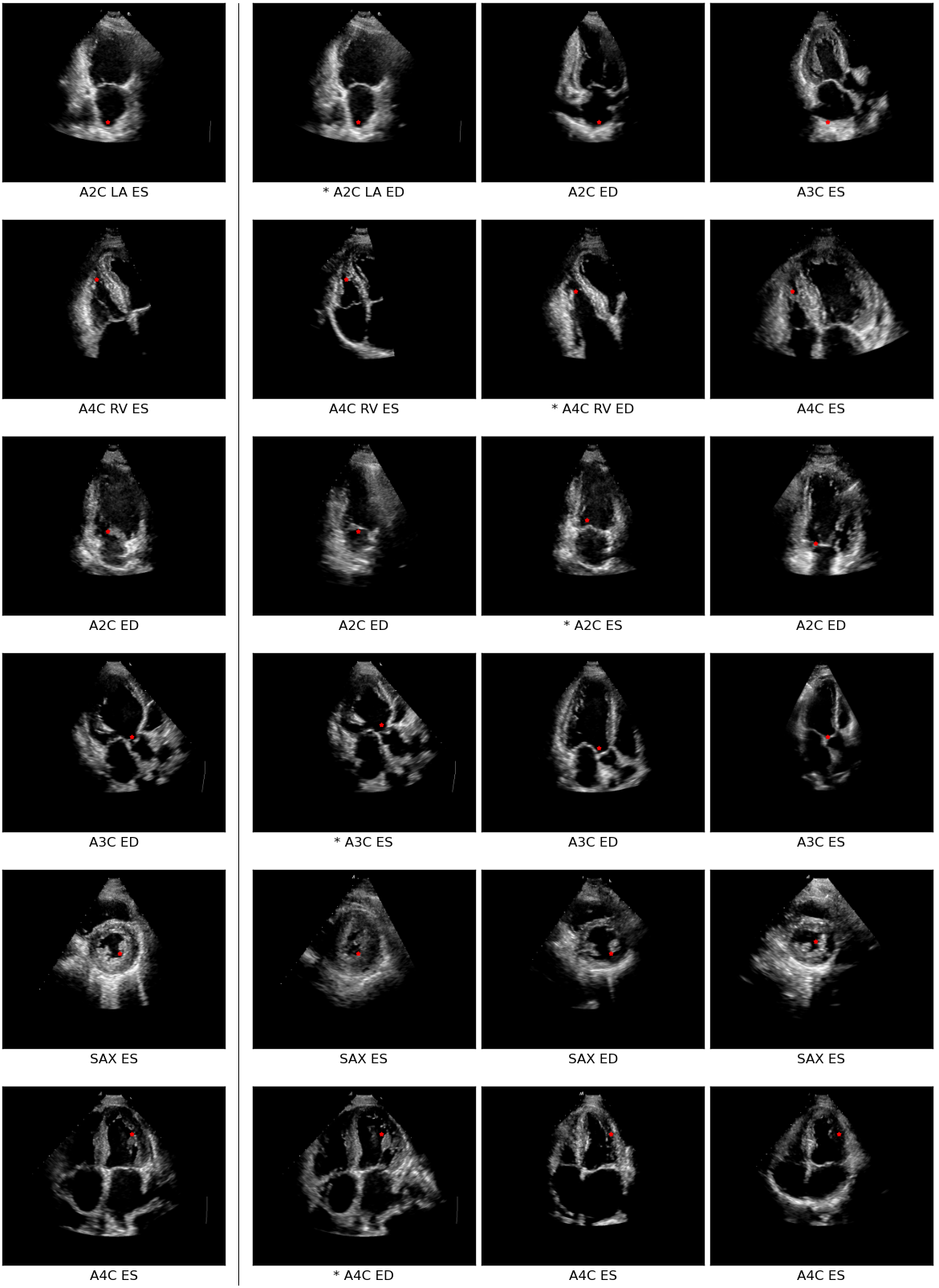}
	\caption{\hl{Region-based} 
 instance retrieval using conceptual search. The~leftmost column displays query images, while the last three columns show the top-3 kNN retrieval results. Red dots indicate the centers of the query and matched descriptor regions. Below~each image, the~view and cardiac phase are displayed. Matches marked with an asterisk (*) are from the same acquisition as the query image, but~from a different cardiac~phase.}
	\label{retr_examples}
\end{figure}

For instance retrieval, nearest-neighbor matching based on the Euclidean distance between descriptors can be employed. 
Initially, we conduct a qualitative analysis by randomly sampling images from the test set and manually selecting specific query points to analyze the results.
The descriptors corresponding to these selected query points were then used to search the database and retrieve samples with regions similar to the query points.
Figure~\ref{retr_examples} showcases six randomly sampled examples, which illustrate that the retrieved image regions align well with the query semantics. 
For example, the~retrieved regions share the same cardiac chamber and view as the query points. 
Moreover, the~anatomical structures around the matched locations are visually similar to those in the query~points.

For the retrieval task, the~search is based solely on the concept descriptors
This approach ensures that the retrieval process focuses on the semantic content rather than stylistic~variations.

To quantitatively analyse this task, we use an independent test set of 450 images, totalling 113,400 region descriptors ($14\cdot18\cdot450$).
Performing nearest neighbor search on this space is very fast. 
The set includes four echocardiographic views (apical 2-, 3-, and~4-chamber views, and~a short-axis view), with~frames captured at end-diastole (ED) and end-systole (ES).
For the apical views, \ac{LV} contour annotations were available, from~which we extracted five key landmark points: left and right annulus, apex, mid-septum, and~mid-free-wall. 
We exploit these annotations to setup a retrieval tasks for these landmark points.
In total, there were 150 ED apical frames, each with five locations used as query points. 
The search pool consisted of all 225 ES frames from all views, including the short-axis view. 
A retrieval is considered a match if it corresponds to the ES image of the ED query and if the retrieved location is adjacent to the annotated landmark~point.

We present the results in Table~\ref{tab_retrieval}, which shows the \ac{mAP} metrics for both models,  computed using the top-5 search results. 
We observe that ConceptVAE demonstrates more than double the performance of the baseline without any retraining, revealing two important observations about ConceptVAE:

\begin{itemize}
	\item The concept probability grid indeed encodes semantic content and thus $x_{concept}$ functions as a spatial arrangement of concepts, which for ConceptVAE are defined as composable higher-level discrete features.
	\item ConceptVAE shows promising results for zero-shot instance retrieval based on local-region queries, unlike more traditional approaches that operate at the image level and need additional fine-tuning.
\end{itemize}


\begin{table}[H]
    \caption{Region-based instance retrieval mAP metric~values.}
    \label{tab_retrieval}
    \centering
     
   \begin{tabularx}{0.4\textwidth}{ccc}
\toprule
         & \multicolumn{2}{ c }{\textbf{\hl{Model}}}  
\\
         \cmidrule{2-3}
        \textbf{\hl{Landmark}} & \textbf{ConceptVAE} & \textbf{Baseline} \\
        \midrule
        left annulus & 0.418 & 0.148 \\
        mid-septum & 0.281 & 0.098 \\
        apex & 0.518 & 0.345 \\
        mid-free-wall & 0.263 & 0.094 \\
        right annulus & 0.371 & 0.128 \\
        \textit{\hl{average}} & 0.370 & 0.163 \\
\bottomrule
    \end{tabularx}
\end{table}

\subsection{Semantic~Segmentation}
\label{subsec_semantic_segm}

The second task we employ is semantic segmentation, where features from the pre-trained models are projected to match a down-sampled ground-truth mask.
For this task, we use five labels corresponding to heart chambers: left and right ventricles and atria in apical views (A2C, A3C, and~A4C views) and the left ventricle in the short-axis (SAX) view.

Starting with frozen model latent codes, a~linear 2D convolutional kernel is fitted to predict low-resolution (stride 16$\times$) segmentation maps. 
Channel-wise softmax activation is applied on top of the predicted linear logits, as~shown in Equation~(\ref{linear_segm}). 
Here,  $p_{ij}(s)$ represents the probability that location $i,j$ to contain chamber $s$, $x_{input}$ is the frozen latent feature map, and~$W_k$ and $w_b$ are the kernel weight matrix and bias vector, respectively, and~containing 6 rows for the 5 prediction targets and one background channel.
\begin{equation}
	\label{linear_segm}
	p_{ij}(s) = Softmax(W_k\cdot x_{input}^{ij} + w_b)
\end{equation}

The ground-truth was obtained by down-sampling the full scale chamber masks using the area interpolation method.
We perform training on an independent set consisting of 5000 training examples, and~test the outcomes using an independent test set of 500 samples. 
The Dice loss was employed as in Equation~(\ref{linear_dice}), where $p_{ij}$ and $t_{ij}$ are the predicted and target chamber presence probabilities at location $i, j$, respectively.
\begin{equation}
	\label{linear_dice}
	\mathcal{L}_{Dice} = 1-\frac{2\sum_{i,j} p_{ij}t_{ij}}{\sum_{i,j} p_{ij}^2+t_{ij}^2}
\end{equation}

We explore three scenarios:  (i) using only the concepts $x_{concept}$ as input, (ii) using the full latent space ($x_{latent} = [x_{concept}, x_{style}]$) as input, and~(iii) using only the style map $x_{style}$ as input.
We also investigate the influence of the linear kernel spatial size $k$ for the \emph{Feature decoder} block on the evaluation scores, with~different ranges, $k \in \{1,3,5,7,9\}$. 
To investigate the effect of the proposed training procedure, we first compare with a randomly initialized frozen model. 
The same random seed, dataset and number of linear-classifier optimization iterations were used throughout all~scenarios. 

Table~\ref{tab_dice} presents the linear evaluation results in terms of Dice Loss, which is equivalent to subtracting the Dice Score from 1. 
For both types of models (trained and randomly initialized) and across all $x_{input}$ setups, larger values of $k$ result in lower test set losses. 
This is expected, as~larger kernels capture more local information, and~concepts cooperate locally to form larger anatomical structures. 
When  $x_{input}:=x_{latent}$ and the model is trained, the~loss decreases only marginally when $k$ exceeds 5 (\ie~the receptive field size used in the \emph{Feature Decoder} block).

In all scenarios, ConceptVAE achieves lower test losses. 
For both models, the~lowest losses occur when  $x_{input}:=x_{latent}$ (\ie both concepts and styles are used for segmentation). 
When using only the concepts from the trained model, the~losses are slightly higher but still significantly lower than when using only styles. 
Additionally, when  $x_{input}:=x_{styles}$, the~differences between the ConceptVAE and the random-init model are the smallest among all three input scenarios. 
This result brings further evidence that  $x_{concept}$ contains semantic information useful for downstream tasks like segmentation, while  $x_{style}$ focuses on local stylistic features. 
Moreover, there are virtually no differences in losses between using only $x_{concept}$ or only $x_{style}$ for the randomly initialised model, whereas these two scenarios yield substantial differences for ConceptVAE. 
This highlights the impact of our proposed unsupervised training framework on the model's ability to separate concepts from~styles.

We also evaluate against the Vicreg baseline model using a similar procedure, but~only for the 1 $\times$ 1 sized convolutional kernel (details provided in Section~\ref{sec_model_analysis}), and~illustrate the outcomes in Table~\ref{tab_dice}. 
We note that ConceptVAE, using trained concepts and 5 $\times$ 5 windows or larger, achieves superior Dice metrics. 
This highlights the benefits of content-style disentanglement and the model's robustness against feature~collapse.

\begin{table}[H]
	\caption{ Dice loss on the semantic segmentation test set when using $x_{concept}$ only, $x_{style}$ only, or~$x_{concept}$ along with $x_{style}$. For each row, the lowest Dice losses are marked with bold.}
	\centering
	 \begin{tabularx}{0.65\textwidth}{ccccc}
\toprule
	& \textbf{Kernel} & \textbf{Concept Only} &  \textbf{Style Only} & \textbf{Concept \& Style} \tabularnewline
	\midrule 
	\multirow{6}{3.5em}{\centering \textbf{\hl{C}oncept VAE}}  
 & 1 $\times$ 1 & $0.5876$  & $0.6641$ & \textbf{0.4853}  \\
 
	\cmidrule{2-5} 
	& 3 $\times$ 3 & $0.2268$ & $0.4238$ & \textbf{0.1741}  \tabularnewline
	\cmidrule{2-5} 
	& 5 $\times$ 5 & $0.1311$ & $0.2586$  & \textbf{0.1087} \tabularnewline
	\cmidrule{2-5}
	& 7 $\times$ 7 & $0.1013$ & $0.1825$ & \textbf{0.0938}  \tabularnewline
	\cmidrule{2-5}
	& 9 $\times$ 9 & $0.0903$ & $0.1520$  & \textbf{0.0900} \tabularnewline
	\midrule 
	\multirow{6}{3.5em}{\centering \textbf{Concept VAE Rand. init.}} & 1 $\times$ 1 & $0.6958$  & $0.6942$ & \textbf{0.6790} \tabularnewline
	\cmidrule{2-5} 
	& 3 $\times$ 3 & $0.5413$ & $0.5205$  & \textbf{0.4655} \tabularnewline
	\cmidrule{2-5}
	& 5 $\times$ 5 & $0.3665$ & $0.3504$  & \textbf{0.2901}  \tabularnewline
	\cmidrule{2-5}
	& 7 $\times$ 7 & $0.2465$ & $0.2405$ & \textbf{0.2016}  \tabularnewline
	\cmidrule{2-5}
	& 9 $\times$ 9 & $0.1876$ & $0.1990$  & \textbf{0.1715}  \tabularnewline
	\midrule

	\textbf{Vicreg} & 1 $\times$ 1 &  \multicolumn{2}{c}{0.187} \tabularnewline
    \bottomrule
 
\end{tabularx}
\label{tab_dice}
\end{table}

\subsection{Near-OOD~Detection}

To assess the proposed model's capability to detect \ac{OOD} samples, we employed a test set comprising only parasternal long-axis (PLAX) views. 
Unlike the test set from Section~\ref{ss_inst_retrieval}, which includes only apical and short-axis acquisitions, this set is considered OOD because, although~it contains echocardiographies, the~views are different. 
The aim of this analysis is to determine whether the latent space features can differentiate between the two data distributions (\ie apical and SAX versus PLAX views).

Most \ac{OOD} methods are designed to work with supervised classification models \mbox{(\eg~\cite{ood_maha,ood_knn})}, thus requiring explicit labeling either for in-domain classes or for flagging outlier samples.
One method that does not require any labels and allows for fast log-likelihood evaluation with respect to the underlying data distribution is \acp{NF}.
To this end, linear \acp{NF}~\cite{nf_survey} were fitted solely on the frozen embeddings of in-distribution data (\ie apical and SAX views) for both the proposed and baseline models. The~\ac{NF} took the form of Equation~(\ref{eq_nf}), where $x$ represents an input derived from the latent space, $y$ is the transformed variable, and~$A, b$ are trainable parameters.
\begin{equation}
    \begin{aligned}
        y &= Ax+b \\
        \ln{p(x)} &= \ln{p_{prior}(y)} + \ln{|\det{A}|} \\
        p_{prior}(y) &= \mathcal{N}(y | \textbf{0}, I) 
    \end{aligned}
    \label{eq_nf}
\end{equation}

For ConceptVAE, $x$ is formed by concatenating a 5 $\times$ 5 window of concept probabilities, excluding the style component. 
For the baseline model, $x$ is the feature embedding of a single location from the latent space feature grid. 
For all spatial locations corresponding to ultrasound cones within the latent space grid, and~for all training data, the~region descriptors $x$ were extracted and fed into the NF to maximize $\ln{p(x)}$ for in-distribution data. 
The same training data as in Section~\ref{subsec_semantic_segm} was used to fit the \acp{NF} (\ie only apical and SAX views). After~the \acp{NF} converged, an~image-level score was computed for each test sample by averaging the  $\ln{p(x)}$ scores for all grid locations pertaining to the ultrasound~cone.

\begin{figure}[b]
     \centering
    \includegraphics[width=0.55\linewidth]{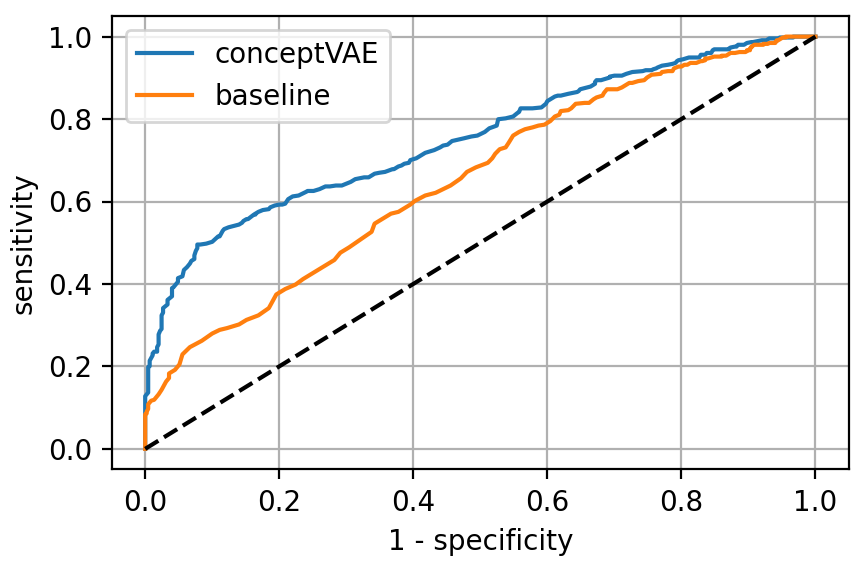}
    \caption{The ROC curves comparison between ConceptVAE and the Vicreg baseline model, for~distinguishing in-distribution echocardiographic views from \ac{OOD} PLAX ones. ConceptVAE has an AuROC score of 0.\hl{753},   
 while the Vicreg baseline has an AuROC of $0.655$.}
    \label{fig_plax_ood}
\end{figure}

Two sets of image-level scores were computed, one for in-distribution apical and SAX views and one for \ac{OOD} PLAX views.
ROC curves were used to assess the score separability between the two sets using ConceptVAE and the Vicreg baseline (Figure \ref{fig_plax_ood}). 
ConceptVAE has an area-under-curve of 0.753, being $10\%$ larger than the baseline (with 0.655). 
 
In contrast to the proposed ConceptVAE, the~baseline model had access to PLAX data during its development (as we used a vast collection of many echocardiography types to pretrain the baseline model, following common practices for classical self-supervized pretraining regarding dataset sizes and variability
, therefore the PLAX view is not \ac{OOD} for the baseline model. Also, the~contrastive objective used for developing the baseline model should promote feature clustering \wrt data sub-groups (\eg anatomical views). Despite this fact, ConceptVAE produces local embeddings that are more separable between echocardiographic views (even \emph{near-OOD} ones), again indicating a reduction of feature collapse due to the content-style disentanglement. This behavior of embeddings separability even for near-OOD data does not usually manifest for regular deep-neural networks~\cite{due}.

\unskip

\subsection{Aortic Valve~Detection}

To further evaluate the generalization capability of ConceptVAE, we aim to detect latent space grid locations corresponding to the aortic valve (AV) region in views not used during pre-training (\ie PLAX).
Similarly to Section~\ref{subsec_semantic_segm}, for~this task we train a linear convolutional layer on top of frozen embeddings to perform a proxy object detection task. 
Each testing sample has a bounding box annotation around the AV along with a label indicating if it’s open or closed (depending on the cardiac phase depicted in the test image). 
We downsized the bounding boxes to the output stride of the latent space and used an overlap threshold
$t$ to determine the \emph{objectness}~\cite{girshick2015fast} of each latent space grid location, \ie~if the down-sampled bounding box overlaps a grid location with a ratio larger than $t$, then that grid location objectness is set as 1, otherwise 0. Moreover, for~each object grid location the newly added convolutional layer also predicts the AV state (open or closed).

For ConceptVAE, the~input to the linear layer is a 5 $\times$ 5 window of both concept probabilities and associated styles for the concepts having the highest probability. 
The output consists of 3 channels, one for classifying objectness
and the other two for classifying the AV state. 
For the baseline Vicreg model, the~setup is similar, but~the input  is the feature vector of a  1x1 latent space grid location (see Section~\ref{sec_model_analysis} for details). 
Balanced binary cross-entropy losses are employed to train both objectives (\ie~detection and labeling).

The results are illustrated in Table~\ref{tab_objDet}.
The mAP scores are close (with the baseline slightly better by $1.6\%$ mAP), while the objectiveness AP is much larger for our proposed model ($+12\%$). This is because our model does a better job in locating Aortic Valve grid positions, but~somewhat lags in correctly classifying the AV state for the detected AV locations. We hypothesise that locating the AV can be done by analyzing concepts (e.g., exploiting a linear separability of concept probabilities w.r.t. AV presence) while the AV state can be inferred from the style component of the latent space. To~test this, we trained a new linear layer only on the concept components of the latent space and observe a severe degradation in label classification performance while retaining the objectness classification performance. The~previous section revealed that the detected concepts on the \textit{near-OoD} PLAX views are still descriptive of the image's semantics; however, the~style component may not fully capture all relevant fine details, since the proposed model was not trained on PLAX views as opposed to the baseline~model.
\begin{table}[H]
    \caption{Mean average precision scores for object detection on PLAX~views.}
    \label{tab_objDet}
    \centering
     
  \begin{tabularx}{0.47\textwidth}{ccc}
\toprule
         & \multicolumn{2}{c}{\textbf{\hl{Model}}}  
 \\
         \cmidrule{2-3} 
         \textbf{\hl{Metric}} & \textbf{ConceptVAE} & \textbf{Baseline} \\
         \midrule
         “open-AV” class AP & 0.337 & 0.297 \\
         “closed-AV” class AP & 0.386 & 0.459 \\
         mean AP & 0.362 & 0.378 \\
         \midrule 
         objectness AP & 0.786 & 0.665 \\
\bottomrule
    \end{tabularx}
\end{table}

\subsection{Style-Based Synthetic Data~Generation}
\label{subsec_style_augm}

We further explore how style information can be used to generate synthetic data.
Such data can be valuable for creating inputs conditioned by patient attributes, such as generating images with more textured walls. 
To achieve this, we leverage the known range of $x_{style}$ (since the constraint 
$\mathcal{L}_{style}$ is enforced during training), and~investigate style-based image generation. 
This involves adding Gaussian noise at various levels 
 $\beta$ as described in Equation~(\ref{style_noise}):
\begin{equation}
	\begin{aligned}
		n &\sim \mathcal{N}(\mathbf{0}, I)\\
		x_{style}^* &= \frac{x_{style}+\beta n}{\sqrt{1+\beta^2}}
	\end{aligned}
	\label{style_noise}
\end{equation}
where $\beta$ controls the amount of noise injected into $x_{style}^*$.

\begin{figure}
	 \centering
	\includegraphics[width=0.99\linewidth]{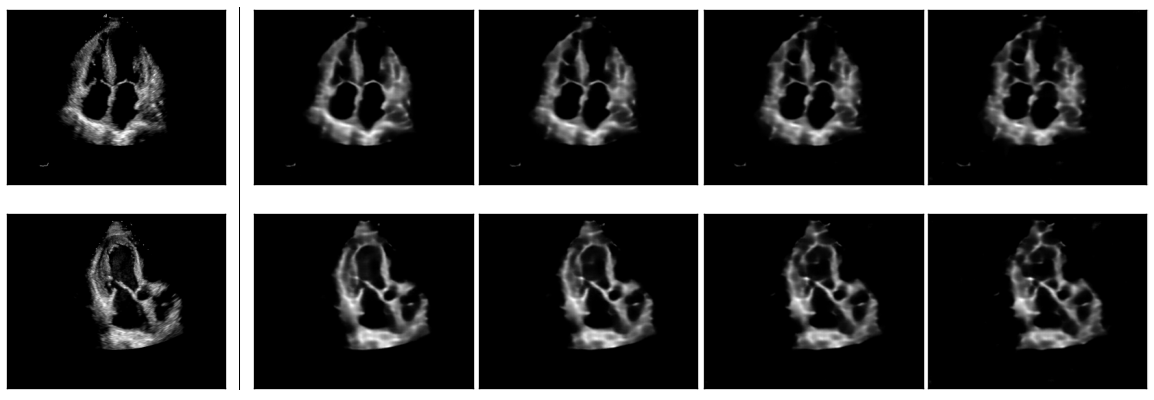}
	\caption{ Original images (left) displayed alongside reconstructions using   $x_{style}^*$ with increasing levels of injected noise, $\beta$. From~the second column to the right, $\beta$ values are $0$ (unaltered reconstruction), $0.2$, $0.4$ and $0.6$, respectively.}
	\label{noise_level}
\end{figure}

\begin{figure}
	 \centering
	\includegraphics[width=0.8\linewidth]{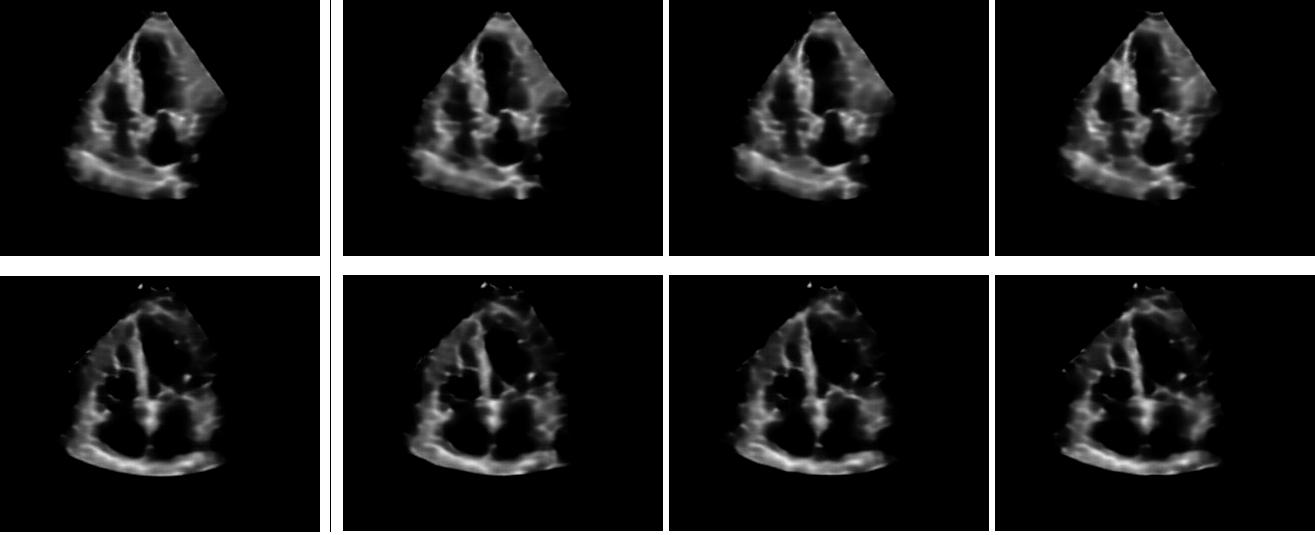}
	\caption{ Reconstructed images with unaltered $x_{style}$ (left) alongside three reconstructions with constant noise level {$\beta=0.3$}. Each noisy reconstruction uses different noise, $n\sim\mathcal{N}(\mathbf{0},I)$, {as described in Equation~(\ref{style_noise})}.}
	\label{noised_samples}
\end{figure}

We then reconstruct the image using these style attributes.
Randomly sampled reconstructions w.r.t.~multiple $\beta$ (reusing the same sampled $n$) are illustrated in Figure~\ref{noise_level}, while Figure~\ref{noised_samples} illustrates reconstructions with multiple noise samplings $n_k \sim \mathcal{N}(\mathbf{0}, I)$ and fixed $\beta=0.3$. 
We observe that even with relatively high $\beta$ values, the~reconstructions closely resemble the unaltered concepts, while the image textures are modified (with minimal changes to anatomical structures in terms of their shape or placement). 
This leads to the following observations:

\begin{itemize}
	\item 
 The model uses  $x_{concept}$ to decode semantic content, such as anatomical structures like chamber walls, blood pools, and~valves, while 
 $x_{style}$ is used  to particularize local textures, shadows and~speckles.
 
	\item With ConceptVAE, synthetic data can be generated by modifying only textures and speckles while retaining anatomical structures. This allows for the generation of novel samples that can serve as style augmentations without modifying the content, potentially enhancing the training performance of dense downstream models, such as those used for segmentation.
\end{itemize}

The samples generated with ConceptVAE remain within the original data distribution, and~thus can serve as a more calibrated augmentation method. 
In contrast, classical transformations such as rotations and blurring may generate data points with appearances not observed in the initial distribution (\eg~unnatural rotations or texture changes). 
Ultrasound medical imaging inherently introduces noise in video acquisitions in the form of pixel speckles.
ConceptVAE simulates the effect of different realizations of echocardiography-specific noise, producing images that reflect this variability. 
Given the large variability between acquisitions and patients in ultrasound imaging~\cite{USvariability}, the~proposed method can potentially improve the robustness of the models on downstream tasks.

\section{Conclusions}
\label{sec:conclusions}

We present ConceptVAE, a~novel \ac{SSL} framework designed to learn disentangled representations of 2D cardiac ultrasound images. 
This method involves converting input embeddings into a set of discrete concepts and associated continuous~styles.

Through multiple qualitative and quantitative analyses, we demonstrate that ConceptVAE captures anatomical information within concepts vectors and local textures within the style vectors, thereby achieving disentanglement. 
For example, by~qualitatively analysing the concept maps, we observe the method is able to specialise certain concepts to independent anatomical structures such as blood pools or septum~walls.

These properties prove beneficial for several downstream applications, including region-based instance retrieval, object detection, and~synthetic data~generation.

Specifically, we provide empirical evidence that ConceptVAE outperforms traditional SSL methods like Vicreg in region-based instance retrieval, \ac{OOD} detection, semantic segmentation, and~object detection. 
Moreover, the~method shows promising results in generating synthetic data samples that reflect the original data distribution and preserve anatomical concepts while varying~styles.

For future work, we propose to apply the method to a broader range of medical image modalities. 
Currently, we evaluated ConceptVAE on cardiac echocardiographies due to the availability of an extensive dataset for pre-training and testing across various downstream tasks.
Additionally, we plan to devise an automated method to identify the number of concepts needed, similar to the way object detection algorithms propose the number of objects present in the image.
Furthermore, we plan to test and extend our method to 3D data, which is prevalent in medical imaging, but~adds another level of complexity both for pre-training and for concept identification. 
In-depth analyses of disentangled representations may also reveal other properties such as enhanced interpretability and explainability, opening promising avenues for future research.
 
\vspace{6pt} 


\textbf{Acknowledgments: }{The 	data used for the empirical experiments are courtesy of Princeton Radiology and Zwanger Pesiri. 
The concepts and information presented in this paper are based on research results that are not commercially available. 
Future commercial availability cannot be~guaranteed.}

\textbf{Author contributions: }{Conceptualization, C.C. and A.S.; methodology, C.C. and A.S.; software, C.C.; validation, C.C.; formal analysis, C.C.; investigation, C.C. and A.S.; resources, P.T.; data curation, P.T.; writing---original draft preparation, C.C.; writing---review and editing, A.S. and P.T.; visualization, C.C.; supervision, A.S.; project administration, P.T.; funding acquisition, P.T.} 


\begin{thebibliography}{99}

\bibitem{taleb20203d}
Taleb, A.; Loetzsch, W.; Danz, N.; Severin, J.; Gaertner, T.; Bergner, B.;
  Lippert, C.
\newblock 3d self-supervised methods for medical imaging.
\newblock {\em Adv. Neural Inf. Process. Syst.} {\bf 2020},
  {\em 33},~18158--18172.

\bibitem{azizi2021big}
Azizi, S.; Mustafa, B.; Ryan, F.; Beaver, Z.; Freyberg, J.; Deaton, J.; Loh,
  A.; Karthikesalingam, A.; Kornblith, S.; Chen, T.;  et~al.
\newblock Big self-supervised models advance medical image classification.
\newblock In Proceedings of the  IEEE/CVF International
  Conference on Computer Vision, \hl{Montreal, BC, Canada, 11--17 October} 
 2021; pp. 3478--3488.

\bibitem{huang2024systematic}
Huang, Z.; Jiang, R.; Aeron, S.; Hughes, M.C.
\newblock Systematic comparison of semi-supervised and self-supervised learning
  for medical image classification. 
\newblock In Proceedings of the  IEEE/CVF Conference on
  Computer Vision and Pattern Recognition, \hl{Seattle, WA, USA, 16--22 June } 2024; pp. 22282--22293.

\bibitem{ssl_cookbook}
Balestriero, R.; Ibrahim, M.; Sobal, V.; Morcos, A.; Shekhar, S.; Goldstein,
  T.; Bordes, F.; Bardes, A.; Mialon, G.; Tian, Y.;  et~al.
\newblock A Cookbook of Self-Supervised Learning.
\newblock {\em arXiv} {\bf 2023}.
   {http://arxiv.org/abs/2304.12210}.

\bibitem{cabannes2023ssl}
Cabannes, V.; Kiani, B.; Balestriero, R.; LeCun, Y.; Bietti, A.
\newblock The ssl interplay: Augmentations, inductive bias, and generalization.
\newblock In Proceedings of the International Conference on Machine Learning,
  PMLR, \hl{ Honolulu, HI, USA, 23--29 July} 2023; pp. 3252--3298.

\bibitem{wu2024voco}
Wu, L.; Zhuang, J.; Chen, H.
\newblock Voco: A simple-yet-effective volume contrastive learning framework
  for 3d medical image analysis.
\newblock In Proceedings of the  IEEE/CVF Conference on
  Computer Vision and Pattern Recognition, \hl{Seattle, WA, USA, 16--22 June } 2024; pp. 22873--22882.

\bibitem{baevski2022data2vec}
Baevski, A.; Hsu, W.N.; Xu, Q.; Babu, A.; Gu, J.; Auli, M.
\newblock Data2vec: A general framework for self-supervised learning in speech,
  vision and language.
\newblock In Proceedings of the International Conference on Machine Learning,
  PMLR,  \hl{Baltimore, MD, USA, 25-27 July} 2022; pp. 1298--1312.

\bibitem{wang2023swinmm}
Wang, Y.; Li, Z.; Mei, J.; Wei, Z.; Liu, L.; Wang, C.; Sang, S.; Yuille, A.L.;
  Xie, C.; Zhou, Y.
\newblock Swinmm: Masked multi-view with swin transformers for 3d medical image
  segmentation.
\newblock In Proceedings of the International Conference on Medical Image
  Computing and Computer-Assisted Intervention, \hl{Vancouver, BA, Canada, 8--12 October 2023;} Springer:  \hl{Berlin/Heidelberg, Germany,} 
  2023; pp. 486--496.

\bibitem{liu2024benchmarking}
Liu, P.; Zhang, J.; Wu, X.; Liu, S.; Wang, Y.; Feng, L.; Diao, Y.; Liu, Z.;
  Lyu, G.; Chen, Y.
\newblock Benchmarking Supervised and Self-Supervised Learning Methods in A
  Large Ultrasound Multi-task Images Dataset.
\newblock {\em IEEE J. Biomed. Health Inform.} {2024}, \emph{\hl{Early Access}}.

\bibitem{holste2024efficient}
Holste, G.; Oikonomou, E.K.; Mortazavi, B.J.; Wang, Z.; Khera, R.
\newblock Efficient deep learning-based automated diagnosis from
  echocardiography with contrastive self-supervised learning.
\newblock {\em Commun. Med.} {\bf 2024}, {\em 4},~133.

\bibitem{insclr}
Deng, Z.; Zhong, Y.; Guo, S.; Huang, W.
\newblock InsCLR: Improving Instance Retrieval with Self-Supervision.
\newblock {\em Proc. AAAI Conf. Artif. Intell.} {\bf 2022},
  {\em 36},~516--524.
\newblock {\url{https://doi.org/10.1609/aaai.v36i1.19930}}.

\bibitem{retrieval_survey}
Chen, W.; Liu, Y.; Wang, W.; Bakker, E.M.; Georgiou, T.; Fieguth, P.; Liu, L.;
  Lew, M.S.
\newblock Deep Learning for Instance Retrieval: A Survey.
\newblock {\em IEEE Trans. Pattern Anal. Mach. Intell.} {\bf
  2023}, {\em 45},~7270--7292.
\newblock {\url{https://doi.org/10.1109/TPAMI.2022.3218591}}.

\bibitem{human_psych}
Bracci, S.; Op~de Beeck, H.
\newblock Understanding Human Object Vision: A Picture Is Worth a Thousand
  Representations.
\newblock {\em Annu. Rev. Psychol.} {\bf 2023}, {\em 74},~113--135.
\newblock
  {\url{https://doi.org/https://doi.org/10.1146/annurev-psych-032720-041031}}.

\bibitem{dicarlo2012does}
DiCarlo, J.J.; Zoccolan, D.; Rust, N.C.
\newblock How does the brain solve visual object recognition?
\newblock {\em Neuron} {\bf 2012}, {\em 73},~415--434.

\bibitem{wardle2020recent}
Wardle, S.G.; Baker, C.I.
\newblock Recent advances in understanding object recognition in the human
  brain: Deep neural networks, temporal dynamics, and context.
\newblock {\em F1000Research} {\bf 2020}, {\em 9}, \hl{590}.

\bibitem{zhang2023dive}
Zhang, C.; Zheng, H.; Gu, Y.
\newblock Dive into the details of self-supervised learning for medical image
  analysis.
\newblock {\em Med Image Anal.} {\bf 2023}, {\em 89},~102879.

\bibitem{eddahmani2023unsupervised}
Eddahmani, I.; Pham, C.H.; Napol{\'e}on, T.; Badoc, I.; Fouefack, J.R.;
  El-Bouz, M.
\newblock Unsupervised learning of disentangled representation via
  auto-encoding: A survey.
\newblock {\em Sensors} {\bf 2023}, {\em 23},~2362.


\bibitem{simclr}
Chen, T.; Kornblith, S.; Norouzi, M.; Hinton, G.
\newblock A simple framework for contrastive learning of visual
  representations.
\newblock In Proceedings of the   37th International Conference on Machine
  Learning, ICML'20, \hl{Virtual, 13--18 July}  2020\hl{.} 


\bibitem{moco}
He, K.; Fan, H.; Wu, Y.; Xie, S.; Girshick, R.
\newblock Momentum Contrast for Unsupervised Visual Representation Learning.
\newblock In Proceedings of the IEEE/CVF Conference on Computer Vision and Pattern
  Recognition (CVPR), \hl{Seattle, WA, USA, 13--19 June} 2020; pp. 9726--9735.
\newblock {\url{https://doi.org/10.1109/CVPR42600.2020.00975}}.

\bibitem{vicreg}
Bardes, A.; Ponce, J.; LeCun, Y.
\newblock {VICR}eg: Variance-Invariance-Covariance Regularization for
  Self-Supervised Learning.
\newblock In Proceedings of the   10th International Conference on Learning
  Representations, \hl{Virtual,  25--29 April} 2022.

\bibitem{masked_autoencoder}
He, K.; Chen, X.; Xie, S.; Li, Y.; Dollár, P.; Girshick, R.
\newblock Masked Autoencoders Are Scalable Vision Learners.
\newblock In Proceedings of the IEEE/CVF Conference on Computer Vision and Pattern
  Recognition (CVPR), \hl{New Orleans, LA, USA, 18--24 June} 2022; \mbox{pp. 15979--15988}.
\newblock {\url{https://doi.org/10.1109/CVPR52688.2022.01553}}.

\bibitem{data2vec}
Baevski, A.; Babu, A.; Hsu, W.N.; Auli, M.
\newblock Efficient Self-supervised Learning with Contextualized Target
  Representations for Vision, Speech and Language.
\newblock In Proceedings of the   40th International Conference  on Machine
  Learning,  ICML'23, \hl{Honolulu, HI, USA, 23--29 July} 2023.

\bibitem{self_duality}
Garrido, Q.; Chen, Y.; Bardes, A.; Najman, L.; LeCun, Y.
\newblock On the duality between contrastive and non-contrastive
  self-supervised learning.
\newblock In Proceedings of the   11th International Conference on Learning
  Representations, \hl{Kigali, Rwanda,  1--5 May} 2023.

\bibitem{vit}
Dosovitskiy, A.; Beyer, L.; Kolesnikov, A.; Weissenborn, D.; Zhai, X.;
  Unterthiner, T.; Dehghani, M.; Minderer, M.; Heigold, G.; Gelly, S.;  et~al.
\newblock An Image is Worth 16x16 Words: Transformers for Image Recognition at
  Scale.
\newblock In Proceedings of the  9th International Conference on Learning
  Representations, \hl{Vienna, Austria, 5 May} 2021.

\bibitem{emerging_props}
Caron, M.; Touvron, H.; Misra, I.; Jegou, H.; Mairal, J.; Bojanowski, P.;
  Joulin, A.
\newblock Emerging Properties in Self-Supervised Vision Transformers.
\newblock In Proceedings of the IEEE/CVF International Conference on Computer Vision (ICCV), \hl{Montreal, BC, Canada, 11--17 October}
  2021; pp. 9630--9640.
\newblock {\url{https://doi.org/10.1109/ICCV48922.2021.00951}}.

\bibitem{tian2024learning}
Tian, Y.; Fan, L.; Chen, K.; Katabi, D.; Krishnan, D.; Isola, P.
\newblock Learning vision from models rivals learning vision from data.
\newblock In Proceedings of the  IEEE/CVF Conference on
  Computer Vision and Pattern Recognition, \hl{Seattle, WA, USA, 16--22 June} 2024; \mbox{pp. 15887--15898}.

\bibitem{acseg}
Li, K.; Wang, Z.; Cheng, Z.; Yu, R.; Zhao, Y.; Song, G.; Liu, C.; Yuan, L.;
  Chen, J.
\newblock ACSeg: Adaptive Conceptualization for Unsupervised Semantic
  Segmentation.
\newblock In Proceedings of the IEEE/CVF Conf. on Computer Vision and Pattern
  Recognition (CVPR), \hl{Vancouver, BC, Canada,  17--24 June} 2023; pp. 7162--7172.
\newblock {\url{https://doi.org/10.1109/CVPR52729.2023.00692}}.

\bibitem{disentangled_1}
Liu, X.; Sanchez, P.; Thermos, S.; O’Neil, A.Q.; Tsaftaris, S.A.
\newblock \hl{Learning disentangled representations in the imaging domain.}
\newblock {\em Med Image Anal.} {\bf 2022}, {\em 80},~102516.
\newblock {\url{https://doi.org/https://doi.org/10.1016/j.media.2022.102516}}.

\bibitem{disentangled_2}
Wang, X.; Chen, H.; Tang, S.; Wu, Z.; Zhu, W.
\newblock Disentangled Representation Learning.
\newblock {\em arXiv} {\bf 2023}.
   {http://arxiv.org/abs/2211.11695}.

\bibitem{locatello2019challenging}
Locatello, F.; Bauer, S.; Lucic, M.; Raetsch, G.; Gelly, S.; Sch{\"o}lkopf, B.;
  Bachem, O.
\newblock Challenging common assumptions in the unsupervised learning of
  disentangled representations.
\newblock In Proceedings of the International Conference on Machine Learning,
  PMLR, \hl{Long Beach, CA, USA, 9--15 June} 2019; pp. 4114--4124.

\bibitem{cluster_gan}
Mukherjee, S.; Asnani, H.; Lin, E.; Kannan, S.
\newblock ClusterGAN: Latent Space Clustering in Generative Adversarial
  Networks.
\newblock {\em Proc. AAAI Conf. Artif. Intell.} {\bf
  2019}, {\em 33},~4610--4617.
\newblock {\url{https://doi.org/10.1609/aaai.v33i01.33014610}}.

\bibitem{style_content_disentang}
Ngweta, L.; Maity, S.; Gittens, A.; Sun, Y.; Yurochkin, M.
\newblock Simple disentanglement of style and content in visual
  representations.
\newblock In Proceedings of the   40th International Conferenceon Machine
  Learning,    ICML'23, \hl{Honolulu, HI,  USA,  23--29 July} 2023\hl{.} 

\bibitem{vqvae2}
Razavi, A.; van~den Oord, A.; Vinyals, O.  Generating diverse high-fidelity
  images with VQ-VAE-2.
\newblock In   Proceedings of the 33rd International Conference on Neural Information Processing
  Systems,  \hl{Vancouver, BC, Canada, 8--14 December 2019}; Curran Associates Inc.: Red Hook, NY, USA,  2019.

\bibitem{taming}
Esser, P.; Rombach, R.; Ommer, B.
\newblock Taming Transformers for High-Resolution Image Synthesis.
\newblock In Proceedings of the IEEE/CVF Conference on Computer Vision and Pattern
  Recognition (CVPR), \hl{Nashville, TN, USA, 20--25 June} 2021; pp. 12868--12878.
\newblock {\url{https://doi.org/10.1109/CVPR46437.2021.01268}}.

\bibitem{chartsias}
Chartsias, A.; Joyce, T.; Papanastasiou, G.; Semple, S.; Williams, M.; Newby,
  D.E.; Dharmakumar, R.; Tsaftaris, S.A.
\newblock Disentangled representation learning in cardiac image analysis.
\newblock {\em Med Image Anal.} {\bf 2019}, {\em 58},~101535.
\newblock {\url{https://doi.org/https://doi.org/10.1016/j.media.2019.101535}}.

\bibitem{wang2023retccl}
Wang, X.; Du, Y.; Yang, S.; Zhang, J.; Wang, M.; Zhang, J.; Yang, W.; Huang,
  J.; Han, X.
\newblock RetCCL: Clustering-guided contrastive learning for whole-slide image
  retrieval.
\newblock {\em Med Image Anal.} {\bf 2023}, {\em 83},~102645.

\bibitem{fischer2023self}
Fischer, M.; Hepp, T.; Gatidis, S.; Yang, B.
\newblock Self-supervised contrastive learning with random walks for medical
  image segmentation with limited annotations.
\newblock {\em Comput. Med Imaging Graph.} {\bf 2023}, {\em
  104},~102174.

\bibitem{gumbel}
Jang, E.; Gu, S.; Poole, B.
\newblock Categorical Reparameterization with Gumbel-Softmax.
\newblock In Proceedings of the  5th  International Conference on Learning
  Representations, \hl{Toulon, France,  24--26 April} 2017.

\bibitem{spade}
Park, T.; Liu, M.; Wang, T.; Zhu, J.
\newblock Semantic Image Synthesis with Spatially-Adaptive Normalization.
\newblock In Proceedings of the IEEE/CVF Conference on Computer Vision and Pattern
  Recognition (CVPR), \hl{Long Beach, CA, USA, 15--20 June} 2019; \mbox{pp. 2332--2341}.
\newblock {\url{https://doi.org/10.1109/CVPR.2019.00244}}.

\bibitem{haghighi2024self}
Haghighi, F.; Taher, M.R.H.; Gotway, M.B.; Liang, J.
\newblock Self-supervised learning for medical image analysis: Discriminative,
  restorative, or adversarial?
\newblock {\em Med Image Anal.} {\bf 2024}, {\em 94},~103086.

\bibitem{RefineNet}
Nekrasov, V.; Shen, C.; Reid, I.
\newblock Light-weight refinenet for real-time semantic segmentation.
\newblock {\em arXiv} {\bf 2018}.
   \url{http://arxiv.org/abs/1810.03272}.

\bibitem{kobayashi2024sketch}
Kobayashi, K.; Gu, L.; Hataya, R.; Mizuno, T.; Miyake, M.; Watanabe, H.;
  Takahashi, M.; Takamizawa, Y.; Yoshida, Y.; Nakamura, S.;  et~al.
\newblock Sketch-based semantic retrieval of medical images.
\newblock {\em Med Image Anal.} {\bf 2024}, {\em 92},~103060.

\bibitem{li2018large}
Li, Z.; Zhang, X.; M{\"u}ller, H.; Zhang, S.
\newblock Large-scale retrieval for medical image analytics: A comprehensive
  review.
\newblock {\em Med Image Anal.} {\bf 2018}, {\em 43},~66--84.

\bibitem{ood_maha}
Ren, J.; Fort, S.; Liu, J.; Roy, A.G.; Padhy, S.; Lakshminarayanan, B.
\newblock A Simple Fix to Mahalanobis Distance for Improving Near-OOD
  Detection.
\newblock {\em arXiv} {\bf 2021}.
  \url{http://arxiv.org/abs/2106.09022}.

\bibitem{ood_knn}
Kuan, J.; Mueller, J.
\newblock Back to the Basics: Revisiting Out-of-Distribution Detection
  Baselines.
\newblock {\em arXiv} {\bf 2022}.
  \url{http://arxiv.org/abs/2207.03061}.

\bibitem{nf_survey}
Kobyzev, I.; Prince, S.J.; Brubaker, M.A.
\newblock Normalizing Flows: An Introduction and Review of Current Methods.
\newblock {\em IEEE Trans. Pattern Anal. Mach. Intell.}
  {\bf 2021}, {\em 43},~3964--3979.
\newblock {\url{https://doi.org/10.1109/TPAMI.2020.2992934}}.

\bibitem{due}
van Amersfoort, J.; Smith, L.; Jesson, A.; Key, O.; Gal, Y.
\newblock On Feature Collapse and Deep Kernel Learning for Single Forward Pass
  Uncertainty.
\newblock {\em arXiv} {\bf 2022}.
  \url{http://arxiv.org/abs/2102.11409}.

\bibitem{girshick2015fast}
Girshick, R.
\newblock Fast r-cnn.
\newblock In Proceedings of the  IEEE International
  Conference on Computer Vision, \hl{Santiago, Chile, 7--13 December} 2015; pp. 1440--1448.

\bibitem{USvariability}
Letnes, J.M.; Eriksen-Volnes, T.; Nes, B.; Wisløff, U.; Salvesen, O.; Dalen,
  H.
\newblock Variability of echocardiographic measures of left ventricular
  diastolic function. The HUNT study.
\newblock {\em Echocardiography} {\bf 2021}, {\em 38},~901--908.
\newblock {\url{https://doi.org/https://doi.org/10.1111/echo.15073}}.

\end{thebibliography}


\end{document}